\documentclass[letterpaper, 10 pt, conference]{ieeeconf}  
\usepackage{booktabs}
\usepackage{graphicx}
\usepackage{booktabs}
\usepackage{amsmath}
\usepackage{array}
\usepackage{lmodern}
\usepackage{ifluatex}
\usepackage{times}
\usepackage{amssymb}
\usepackage{subfigure}
\usepackage{svg}
\usepackage{grffile}
\usepackage{afterpage}
\usepackage{capt-of}
\usepackage{placeins} 
\usepackage{multirow}
\usepackage[colorlinks=true, linkcolor=blue, citecolor=blue, urlcolor=blue ]{hyperref}

\IEEEoverridecommandlockouts                              

\begin{document}

\title{\LARGE \bf
Multi-Type Preference Learning: Empowering Preference-Based Reinforcement Learning with Equal Preferences
}

\author{Ziang Liu$^{1}$,\thanks{$^{1}$School of Computer Science and Technology, East China Normal University, Shanghai, China}
Junjie Xu$^{1}$,
Xingjiao Wu$^{2}$, \thanks{$^{2}$School of Computer Science, Fudan University, Shanghai, China}
Jing Yang$^{1*}$, 
Liang He$^{1*}$ \thanks{\tt\small Mail addresses: 51265901036@stu.ecnu.edu.cn, jjxu\_dr@stu.ecnu.edu.cn, xjwu\_cs@fudan.edu.cn, jyang@cs.ecnu.edu.cn, lhe@cs.ecnu.edu.cn}
}

\maketitle
\begin{abstract} 
Preference-Based reinforcement learning (PBRL) learns directly from the preferences of human teachers regarding agent behaviors without needing meticulously designed reward functions. However, existing PBRL methods often learn primarily from explicit preferences, neglecting the possibility that teachers may choose equal preferences. This neglect may hinder the understanding of the agent regarding the task perspective of the teacher, leading to the loss of important information. To address this issue, we introduce the Equal Preference Learning Task, which optimizes the neural network by promoting similar reward predictions when the behaviors of two agents are labeled as equal preferences. Building on this task, we propose a novel PBRL method, Multi-Type Preference Learning (MTPL), which allows simultaneous learning from equal preferences while leveraging existing methods for learning from explicit preferences. To validate our approach, we design experiments applying MTPL to four existing state-of-the-art baselines across ten locomotion and robotic manipulation tasks in the DeepMind Control Suite. The experimental results indicate that simultaneous learning from both equal and explicit preferences enables the PBRL method to more comprehensively understand the feedback from teachers, thereby enhancing feedback efficiency. 
Project page: \url{https://github.com/FeiCuiLengMMbb/paper_MTPL}
\end{abstract}

\section{INTRODUCTION}

Reinforcement Learning \cite{DRL-2022, MDP-AI, RL-1961} (RL) has demonstrated remarkable accomplishments in solving complex sequential decision-making tasks, such as controlled nuclear fusion \cite{EPFL-Magnetic-control}, drone control \cite{ETH-swift} and
robotic manipulation \cite{DPRL-IJRR,26-icra,27-icra}. 
However, the effectiveness of RL heavily relies on the careful design of reward functions \cite{0-2017survey,23-ICML,mis-specified-1,mis-specified-2}.  
In this context, Preference-Based Reinforcement Learning (PBRL) \cite{0-DRLPB, 19-rlhf, 25-rlhfsurvey, 28-2024human} has emerged as a powerful framework for learning reward functions from human preferences between the behaviors of two different agents, eliminating the need for meticulously designed reward functions.
Despite the successes of PBRL, it typically necessitates a significant amount of teacher feedback, which can be prohibitively expensive.
To enhance feedback efficiency, prior work has proposed several methods, including pre-training strategies \cite{1-PEBBLE}, facilitating agent exploration \cite{2-RUNE}, utilizing unlabeled data \cite{14-loss}, and employing techniques such as data augmentation \cite{3-SURF}, meta-learning \cite{6-FSHITLRL}, bi-level optimization \cite{4-MRnet}, and preference transitivity \cite{11-Seqrank}.
\begin{figure} [htbp]
    \centering
    \includegraphics[scale=0.34]{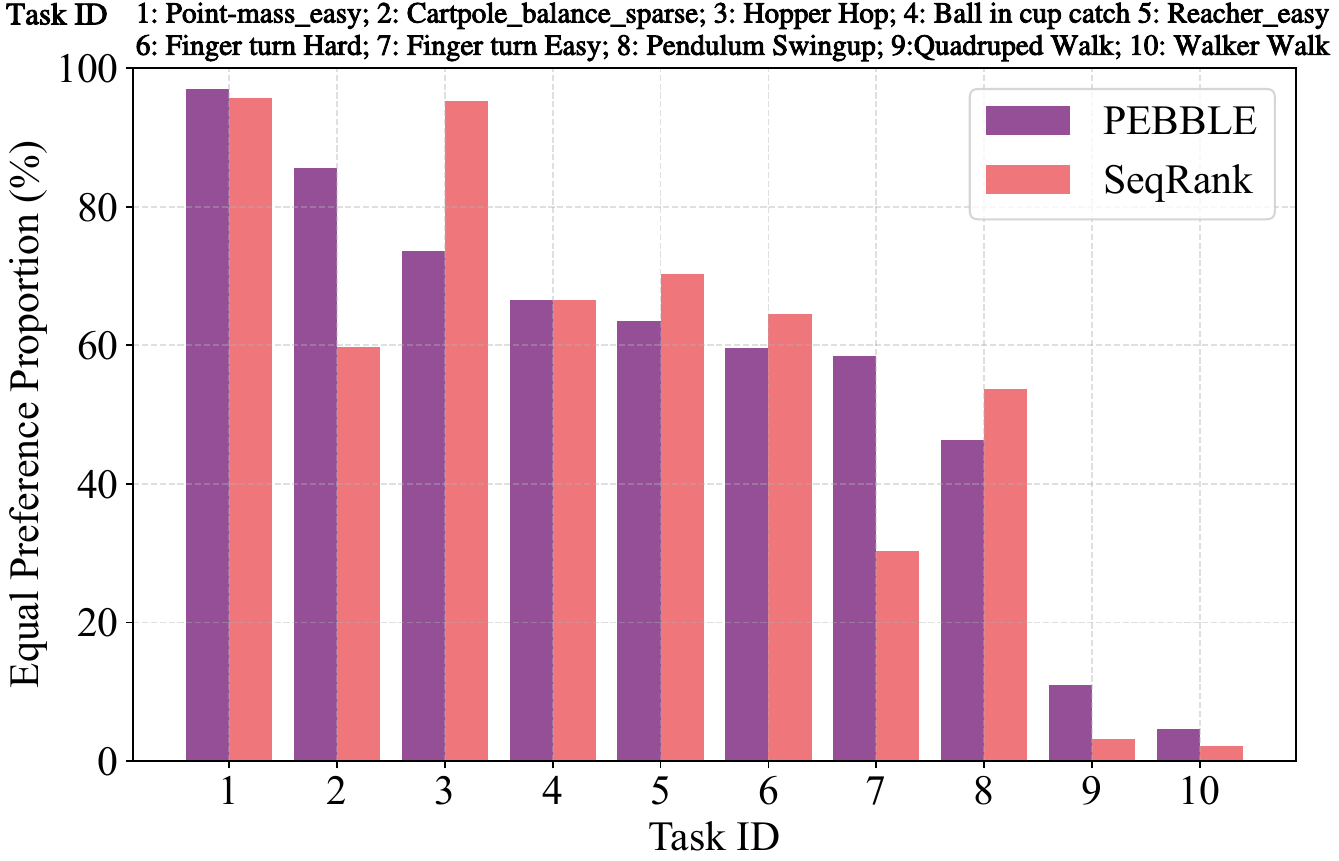}
    \caption{
    Proportion of equal preferences using the Equal SimTeacher \cite{18-B-pre} $(\alpha = 0.1)$ on 10 tasks from the DMC \cite{17-DMC}, tested across two methods \cite{1-PEBBLE,11-Seqrank}. The Equal SimTeacher provides only two types of teacher preferences: equal preferences ($\sigma_0 \simeq \sigma_1$) and explicit preferences ($\sigma_0 \succ \sigma_1$).
    }
    \label{fig_equal_preference_proportion}
\end{figure}
However, existing methods predominantly optimize reward functions using explicit preferences ($\sigma_1 \succ \sigma_0$).
This oversight neglects the possibility that teachers frequently exhibit equal preferences ($\sigma_0 \simeq \sigma_1$) when faced with similar action behaviors.
As illustrated in Fig. \ref{fig_equal_preference_proportion}, there is a significant proportion of equal preferences provided by simulated teachers across various tasks.
This prevalence suggests that addressing equal preferences is essential for enhancing feedback efficiency and improving overall learning performance. Neglecting equal preferences may prevent the agent from grasping the teacher's understanding of the tasks, leading to wasted information.
This study seeks to explore methods that incorporate both explicit and equal preferences, thereby filling a critical gap in the current literature.

In this work, we introduce the Equal Preference Learning Task, which encourages the neural network to produce similar reward value predictions when the behaviors of two agents are labeled as equal preferences. 
This approach enables agents to learn directly from equal preferences. 
Furthermore, inspired by the concept of multi-task learning \cite{12-MTLsurvey}, we propose Multi-Type Preference Learning (MTPL), which learns from both equal preferences and explicit preferences simultaneously. 
This approach enhances the understanding of teacher feedback and improves the feedback efficiency of the PBRL method by treating these two types of learning as two tasks.
Our core contributions are as follows:
\begin{enumerate}
    \item We propose a novel PBRL method, MTPL, which learns simultaneously from both equal and explicit preferences. 
    This method leads to a more comprehensive understanding of human feedback and improved feedback efficiency.
    \item We propose the Equal Preference Learning Task, which empowers agents to learn directly from equal preferences, enabling them to better extract and leverage the information in human feedback.
    \item The experiment results highlight the effectiveness of our approach in leveraging both equal and explicit preferences to enhance performance. 
    Specifically, the average performance across eight tasks increased by \(27.34\%\). 
    Notably, in two tasks with a limited number of explicit preferences, we observe substantial improvements: a $40,490\%$ increase in \(Point\_mass\_easy\) and a \(3,188\%\) increase in \(Hopper\_hop\).
\end{enumerate}

\section{RELATED WORK}\label{sec:related}
\textbf{Preference-Based Reinforcement Learning.}
In the field of Preference-Based Reinforcement Learning (PBRL) \cite{0-DRLPB}, 
a core challenge is to solve a specific task with minimal human feedback.
Prior research primarily enhances feedback efficiency through three types of methods.
First, several studies focus on extracting valuable insights from human preferences \cite{1-PEBBLE,14-loss,3-SURF,28-IROS,34-FTB}. 
PEBBLE \cite{1-PEBBLE} leverages an off-policy algorithm \cite{13-SAC} and pre-training to enhance feedback efficiency.
SeqRank \cite{11-Seqrank} explores the transitivity of human preferences and introduces sequential preference ranking to better exploit the information in human feedback.
Second, another area of research focuses on optimizing reward learning \cite{23-ICML,2-RUNE,4-MRnet,5-LSI,10-iclr,20-AAAI24}.
RUNE \cite{2-RUNE} utilizes the uncertainty in reward function outputs to guide exploration.
MRN \cite{4-MRnet} recognizes the impact of Q-function performance on reward function learning and employs bi-level optimization methods to enhance its effectiveness.
Finally, this category focuses on designing neural network architectures to better model human preference \cite{33-PT,22-AAAI2024}.
PT \cite{33-PT} models human preferences using transformers and introduces a new preference model based on the weighted sum of non-Markovian rewards.
Existing PBRL methods primarily rely on explicit preferences from human teachers, indicating which action sequence is preferred.
Considering the simultaneous selection of explicit and equal preferences by humans more closely reflects real-world scenarios.
Our work aims to facilitate the broader application of PBRL by extracting valuable insights from human preferences,\\
\textbf{Multi-Task Learning.}
Multi-Task Learning (MTL) \cite{12-MTLsurvey,29-MTLsurvey} involves jointly training a model on multiple related tasks to enhance learning efficiency through knowledge sharing. 
Recently, MTL has been successfully applied in various fields, including reinforcement learning \cite{14-aaai,32-MTLrl}, recommendation systems \cite{15-nips,30-MTLre} and drug design \cite{16-nmi,31-MTLdd}.
We apply MTL principles to design our method, MTPL, enabling simultaneous learning from equal and explicit preferences to better align with human feedback.
\section{PRELIMINARIES} \label{sec:Pre}
\textbf{Preference-Based Reinforcement Learning} offers a distinct advantage over traditional reinforcement learning by eliminating the need for a pre-designed reward function, allowing for more flexible and adaptable learning.
PBRL employs feedback from a human teacher on the agent's behaviors to learn a reward function. This learned reward function subsequently guides the agent in completing tasks following human feedback.We represent the agent's behavior as a segment $\sigma$, which encompasses a sequence of states and actions.
For a given pair of segments $(\sigma_0, \sigma_1)$, a human provides feedback by indicating either an explicit preference for one segment over the other or by expressing an equal preference if the difference between them is insignificant. 
The value of \(y\) represents human preference: 0 for \(\sigma_0 \succ \sigma_1\), 1 for \(\sigma_1 \succ \sigma_0\), and 0.5 for equal preference \(\sigma_0 \simeq \sigma_1\).
In this paper, we define $y \in \{0, 1\}$ as an explicit preference, with $y = 0.5$ representing an equal preference.
To incorporate human preferences into deep RL, previous works model the preference predictor using the reward function $\hat{r}_\psi$ as follows:
\begin{equation}
    P_\psi[\sigma^0 \succ \sigma^1] = \frac{exp\sum_{t}\hat{r}_\psi(s_{t}^{0},a_{t}^{0})}{\sum_{i\in\{0,1\}exp\sum_{t}\hat{r}_\psi(s_{t}^{i},a_{t}^{i})}}.
\end{equation}
A significant limitation of this modeling approach is its inability to handle cases of equal preference.
\section{Multi-Type Preference Learning}\label{sec:Meth}
\begin{figure*}[htbp]
    \centering
     \includegraphics[width=1\linewidth]{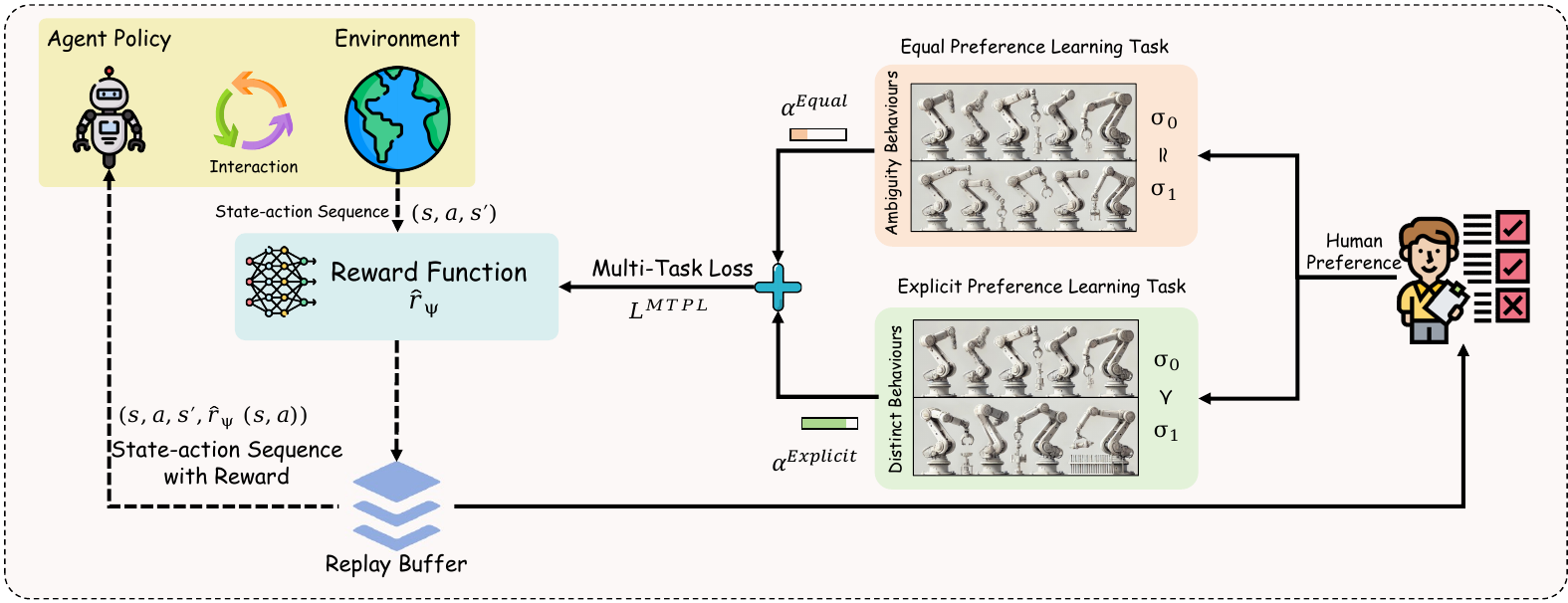}
    \caption{Illustration of MTPL. The agent interacts with the environment and simultaneously learns the reward function $\hat{r}_\psi$ from both equal and explicit preferences.
    State-action sequences are sampled by interacting with the environment, where rewards are labeled by $\hat{r}_\psi$, and transitions are sampled from the replay buffer to optimize the policy.}
    \label{fig_MTPL}
\end{figure*}

In this section, we introduce MTPL: \textbf{M}ulti-\textbf{T}ype \textbf{P}reference \textbf{L}earning. This PBRL method simultaneously learns the reward function from both equal and explicit preferences, treating them as related tasks to enhance overall understanding  (Fig. \ref{fig_MTPL}).
First, we provide a brief overview of the reward learning process from explicit preferences, following established methods in prior work \cite{0-DRLPB,1-PEBBLE}.
Next, we define the Equal Preference Learning Task. 
Finally, we explain how MTPL simultaneously learns from these two related tasks.
\begin{figure}[htbp]
    \centering
    \includegraphics[width=1\linewidth]{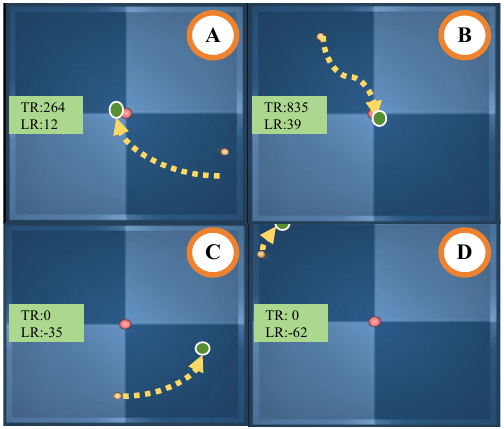}
    \caption{Examples of the behaviors of four agents in the \(Point\_mass\_easy\) task. The objective is for the agent to guide the yellow ball to the central red area. The yellow dashed line represents the agent's path, while the green dot indicates the agent's final stopping position. TR denotes the score given by the true reward function of the simulation environment for the agent's actions, and LR denotes the score from the fitted reward function \(\hat{r}_\psi\).}
    \label{fig_example}
\end{figure}

\subsection{Explicit Preference Learning Task}
In the explicit preference learning task, we use cross-entropy loss to align the preference predictor with the explicit preferences of the (human) teacher.
Specifically, given a dataset of explicit preferences $\mathcal{PPD}$, the reward function, modeled as a neural network with parameters $\psi$, is updated by minimizing the following cross-entropy loss:
\begin{equation}
    \begin{aligned}
    \mathcal{L}^{\text{explicit}} =  - \mathbb{E}_{(\sigma^i,\sigma^j,y)\sim \mathcal{PPD}} [y\log P_\psi[\sigma^i\succ \sigma^j] \\ + (1-y) \log P_\psi[\sigma^j\succ \sigma^i]].
    \end{aligned}
    \label{eq:explicit}
\end{equation}
This task allows the neural network to gain a precise understanding of the behaviors preferred by humans.
We use the $Point\_mass\_easy$ in the DMC as an example (see Fig. \ref{fig_example}).
The behavior of agent B achieves the highest true reward value, making it easy for humans to provide explicit preferences of \(\sigma_B \succ \sigma_{A, C, D}\) compared to the other agents.
These preferences help agents understand that the goal of the $Point\_mass\_easy$ is to reach the central red area.

\subsection{Equal Preference Learning Task}\label{sec:LearningFromEqualPreferences}
In previous PBRL works, the focus has primarily been on learning from explicit preferences, often overlooking equal preferences.
In real-world scenarios, when faced with behaviors that are very similar in task completion, human teachers may choose to express equal preferences.
Therefore, we define the Equal Preference Learning Task to enable PBRL to learn directly from equal preferences.
If two behaviors are labeled as equal preferences, the learned reward function should predict similar reward values for them.
Specifically, we minimize the mean squared error to ensure that the network outputs similar reward values for action sequences labeled as equal preferences ($\sigma_0 \simeq \sigma_1$).
Specifically, we achieves this by minimizing the $\mathcal{L}^{\text{Equal}}$ loss:
\begin{equation}
\mathcal{L}^{\text{Equal}} = \mathbb{E}_{(\sigma^i,\sigma^j)\sim \mathcal{EPD}} [(\hat{r}_\psi(\sigma^i)-\hat{r}_\psi(\sigma^j))^2],
\label{eq:Equal}
\end{equation}
where $\mathcal{EPD}$ represents the dataset of equal preferences.
The output of the neural network for an action sequence $\sigma$, denoted as $\hat{r}_\psi(\sigma)$, models the reward function parameterized by $\psi$.
Optimizing neural networks using an $\mathcal{L}^{\text{Equal}}$ can help them better understand human preferences, even when there are very few explicit preferences.
The task could allow neural networks to learn to distinguish subtle differences in human judgment, thereby more accurately capturing the underlying structure of preferences. 
We also illustrate the Equal Preference Learning Task using the $Point\_mass\_easy$ from the DMC, as shown in Fig. \ref{fig_example}.
Two agents (C and D) all failed to reach the goal. 
Despite their inability to complete the maze, it is extremely challenging for (human) teachers to discern which of the behaviors of the agents is preferable.
Furthermore, for agents C and D, the true reward values representing task completion are both zero. 
However, the output values of the reward function fitted by the neural network are not consistent for these two behaviors.  
This indicates that the current reward function does not recognize that these action sequences are very similar in task completion.
When employing $\mathcal{L}^{\text{Equal}}$, we can enforce the output values of agent behaviors corresponding to equal preference to be as similar as possible. This approach can enable the neural network to more effectively capture human preferences.

By integrating the aforementioned preference learning tasks, we arrive at our final multi-type preference learning method, MTPL, which simultaneously learns from both the Equal and the Explicit preference learning task through linear weighting.
The loss function for MTPL is defined as a weighted sum of the individual task losses:
\begin{equation}
    L^{MTPL} = \alpha^{Explicit} \mathcal{L}^{explicit} + \alpha^{Equal} \mathcal{L}^{Equal},
    \label{eq:MTPL}
\end{equation}
where $\mathcal{L}^{\text{explicit}}$ and $\mathcal{L}^{\text{Equal}}$ represent the losses for the explicit preference task and the equal preference task respectively, and $\alpha^{explicit}$ and $\alpha^{Equal}$ are the corresponding weight hyperparameters.
The agent gains a more comprehensive grasp of human preferences by learning from these two related sub-tasks.
This approach enables the agent to understand not only which behaviors are preferred but also how humans perceive the relative similarity between different behaviors.

\section{EXPERIMENTS}\label{sec:Ex}
In this section, we aim to explore several key aspects of Multi-Type Preference Learning (MTPL).
We design our experiments to answer the following questions:
\begin{itemize}
    \item \textbf{Q1:} Does MTPL improve the feedback efficiency of Preference-Based Reinforcement Learning methods?
    \item \textbf{Q2:} Does simultaneous learning from equal preferences and explicit preferences lead to a better understanding of human feedback?
    \item \textbf{Q3:} Does learning from equal preferences provide meaningful information for task completion?
    \end{itemize}
\subsection{SETUPS}
To evaluate the effectiveness of Mulit-Type Preference Learning (MTPL), we task the agent with solving ten locomotion and robotic manipulation tasks from DeepMind Control Suite (DMC) \cite{17-DMC}.
For simplicity, we use Task ID for representation, with the corresponding task names shown in Fig. \ref{fig_equal_preference_proportion}.
Among these ten tasks, the SimTeacher exhibited a relatively high proportion of equal preferences in eight tasks and a lower proportion in the remaining two (Task ID 9 and 10).
The agent is not accessible to the ground-truth reward function, and all PBRL methods rely on simulated human preference labels.
To better simulate human-like equal preferences, we employed the Equal SimTeacher proposed by B-Pre \cite{18-B-pre}, which provides preferences between two trajectory segments based on the underlying reward function. 
If the difference in the true reward values between two action sequences is less than the threshold \(\delta_{\text{equal}}\), the Equal SimTeacher assigns an equal preference to these trajectories. 
The threshold \(\delta_{\text{equal}}\) is calculated as $\delta_{\text{equal}} = \alpha \frac{\text{Avgret} \times \text{Len(seg)}}{\text{Len(env)}},$
where \(\alpha\) is a hyperparameter called $teacher\_eps\_equal$ , \(\text{Avgret}\) represents the average true return during training, \(\text{Len(seg)}\) denotes the length of the trajectory segment \(\sigma\), and \(\text{Len(env)}\) is the maximum episode length.
In our main experiments, the hyperparameter \(\alpha\) is set to \(0.1\), consistent with the settings used in B-Pre \cite{18-B-pre}.
Figure \ref{fig_equal_preference_proportion} shows the proportion of equal preferences provided by the SimTeacher across these tasks.
To validate the generalizability of MTPL, we apply it to four state-of-the-art baselines: PEBBLE \cite{1-PEBBLE}, RUNE \cite{2-RUNE}, MRN \cite{4-MRnet}, and SeqRank \cite{11-Seqrank}, all of which learn from preferences using cross-entropy loss. 
Since these methods utilize the Soft Actor-Critic (SAC) \cite{13-SAC} algorithm for policy learning, we also compare against SAC with the ground truth reward as an upper bound for our method. Our goal is to closely match SAC's performance using as few preference queries as possible, rather than to outperform it.
\\
\textbf{Implementation details of MTPL.} 
For all experiments, we use the same hyperparameters as in the PEBBLE \cite{1-PEBBLE}. 
We adopt a uniform-based sampling strategy, selecting queries uniformly from the replay buffer, following the setting in \cite{11-Seqrank}.
We set \(\alpha^{\text{Explicit}} = 1.0\).
Detailed settings for $\alpha^{\text{Equal}}$ and the number of feedback are shown in Table \ref{hyperparameters}.
We train the agents using five different seeds and report the average performance with standard deviation (see Table \ref{table_r1}).

\begin{table}[htbp] 
\tabcolsep=0.27cm
    \centering
        \caption{Feedback and $\alpha^{\text{Equal}}$ for each algorithm.} \label{hyperparameters}
    \begin{tabular}{cccccc} 
        \toprule
        Task ID &  feedback & PEBBLE & RUNE& MRN & Seqrank \\
        \midrule
        1       & 0.4K & 0.05  & 0.05  & 0.05 & 0.05\\
        2       & 0.4K & 0.05  & 0.05  & 0.05 & 0.05\\
        3       & 1.0K & 0.05  & 0.05  & 0.05 & 0.05\\
        4       & 0.4K & 0.05  & 0.05  & 0.01 & 0.01\\
        5       & 0.2K & 0.05  & 0.015  & 0.05 & 0.05\\
        6       & 0.4K & 0.05  & 0.05  & 0.05 & 0.04\\
        7       & 0.4K & 0.05  & 0.05  & 0.05 & 0.05\\
        8       & 0.4K & 0.05  & 0.05  & 0.05 & 0.05\\
        9       & 1.0K & 0.05  & 0.01  & 0.05 & 0.05\\
        10      & 0.4K & 0.05  & 0.05  & 0.05 & 0.01\\
    \bottomrule
        
    \end{tabular}
    \end{table}

\subsection{MAIN RESULTS}
\begin{table*}[htbp] 
    \centering
    \caption{
    True Rewards after Convergence in DMC tasks. The Gain ($\%$) column indicates the degree of average improvement achieved by the MTPL method compared with the baseline.}\label{table_r1}
\begin{tabular}{ccrrrrrrrrrr}
\toprule 
Task ID                       &      & 1      & 2      & 3      & 4      & 5      & 6      & 7      & 8      & 9      & 10    \\
\midrule
\multirow{2}{*}{PEBBLE \cite{1-PEBBLE}}       & Mean$\uparrow$  & 1.25           & 584.82                  & 0.17       & 538.10            & 60.18        & 563.12           & 357.46           & 513.48           & 454.33         & 921.54      \\
                              & Std$\downarrow$  & 1.14           & 376.10                  & 0.34       & 288.43            & 18.31        & 386.06           & 170.41           & 322.65           & 260.32         & 56.92       \\
\midrule
\multirow{3}{*}{PEBBLE+MTPL}  & Mean$\uparrow$ & \textbf{677.35}       & \textbf{1000.00}                 & \textbf{20.33}      & \textbf{650.98}            & \textbf{62.80 }       & \textbf{705.92 }          &\textbf{628.94}           & \textbf{535.18 }          & \textbf{559.24}         & \textbf{953.35}      \\
                              & Std$\downarrow$  & 289.99         & 0.00                    & 40.46      & 392.91            & 39.99        & 336.58           & 272.98           & 341.63           & 251.50         & 14.81       \\
                              & Gain ($\%$) & 53915          & 70.99                   & 11860      & 20.98             & 4.35         & 25.36            & 75.95            & 1.98             & 23.09          & 3.45        \\
\midrule
\multirow{2}{*}{RUNE \cite{2-RUNE}}         & Mean$\uparrow$ & 1.14           & 306.34                  & 2.99       & 632.70            & 86.46        & 619.02           & 485.34           & 553.66           & 515.58         & 939.42      \\
                              & Std$\downarrow$  & 0.78           & 364.16                  & 5.41       & 270.31            & 43.60        & 388.33           & 302.47           & 349.84           & 327.38         & 41.41       \\
\midrule
\multirow{3}{*}{RUNE+MTPL}    & Mean$\uparrow$ & \textbf{165.28} & \textbf{835.10} & \textbf{15.45} & \textbf{698.36} & \textbf{102.96} & \textbf{670.16} & \textbf{493.62} & \textbf{649.42} & \textbf{598.11} & \textbf{957.98} \\
                              & Std$\downarrow$  & 289.45         & 329.80                  & 30.31      & 384.05            & 17.03        & 352.15           & 230.61           & 307.43           & 279.48         & 6.03        \\
                              & Gain ($\%$) & 14322          & 172.61                  & 416.75     & 10.38             & 19.08        & 8.26             & 1.71             & 17.30            & 16.00          & 1.98        \\
\midrule
\multirow{2}{*}{MRN \cite{4-MRnet}}          & Mean$\uparrow$ & 1.61           & 589.66                  & 3.86       & 661.66            & 54.76        & 615.41           & 393.58           & 717.30           & 498.39         & 909.86      \\
                              & Std$\downarrow$  & 1.92           & 469.66                  & 5.25       & 366.58            & 13.81        & 305.06           & 248.18           & 142.35           & 232.54         & 73.86       \\
\midrule
\multirow{3}{*}{MRN+MTPL}     & Mean$\uparrow$& \textbf{718.27} & \textbf{807.98} & \textbf{20.38} & \textbf{946.20} & \textbf{97.62} & \textbf{775.44} & \textbf{711.96} & \textbf{753.40} & \textbf{614.90} & \textbf{934.91} \\
                              & Std$\downarrow$  & 183.72         & 384.04                  & 40.68      & 46.55             & 39.89        & 251.07           & 156.86           & 114.48           & 197.44         & 15.39       \\
                              & Gain ($\%$) & 44513          & 37.02                   & 428.30     & 43.00             & 78.27        & 26.00            & 80.89            & 5.03             & 23.38          & 2.75        \\
\midrule
\multirow{2}{*}{SeqRank \cite{11-Seqrank}}      & Mean$\uparrow$ & 0.84         & 681.06                  & 27.78      & 771.96            & 69.94        & 891.94           & 760.24           & 600.44           & 572.53         & 887.68      \\
                              & Std$\downarrow$  & 0.87           & 394.02                  & 54.86      & 305.03            & 38.80        & 109.94           & 189.41           & 258.72           & 343.13         & 111.15      \\
\midrule
\multirow{3}{*}{SeqRank+MTPL} & Mean$\uparrow$ & \textbf{414.21} & \textbf{1000.0} & \textbf{43.41} & \textbf{823.58} & \textbf{83.34} & \textbf{915.44} & \textbf{775.20} & \textbf{672.52} & \textbf{613.04} & \textbf{924.39} \\ 
                              & Std$\downarrow$  & 301.46         & 0.00                    & 52.21      & 186.51            & 62.91        & 67.61            & 249.53           & 141.70           & 308.71         & 67.32       \\
                              & Gain ($\%$) & 49210          & 40.83                   & 50.25      & 6.69              & 19.16        & 2.63            & 1.98             & 12.00            & 7.08           & 4.14        \\
\midrule
\midrule
\multirow{2}{*}{SAC \cite{13-SAC}}          & Mean$\uparrow$ & 897.28         & 899.78                  & 253.81     & 981.70            & 957.70       & 970.64           & 953.32           & 247.90           & 936.22         & 968.84      \\
                              & Std$\downarrow$  & 6.68           & 124.29                  & 37.16      & 4.04              & 36.84        & 11.75            & 37.14            & 303.65           & 10.93          & 4.59        \\
\bottomrule
\end{tabular}
\end{table*}
Results in Table \ref{table_r1} demonstrate the superiority of the proposed MTPL comparison over four SOTA baselines in all tasks, confirming that MTPL effectively enhances the feedback efficiency of PBRL methods (Q1).
We compare the performance of PEBBLE, RUNE, MRN, and SeqRank with and without MTPL respectively.
Overall, the average reward increased by \(27.34\%\) compared to the original baselines, excluding extreme values.
This average reward reflects the mean performance improvement across all tasks.
In task \(Point\_mass\_easy\), where the number of explicit preferences is deficient (less than \(5\%\)), our MTPL method achieves a remarkable average performance improvement of \(40490\%\) compared to the baselines without MTPL. 
Similarly, in task \(Hopper\_hop\), which also exhibits a deficient number of explicit preferences (less than \(27\%\) for PEBBLE and less than \(5\%\) for SeqRank), our MTPL method improves average performance improvement by \(3188\%\) compared to the same baselines.
Notably, even in tasks with a lower proportion of equal preferences (Task 9: $10.96\%$, Task 10: $4.65\%$), our method still achieves significant performance improvements.
This significant boost allows us to successfully solve tasks that previous methods could not complete, highlighting the effectiveness of our MTPL approach.

In addition to demonstrating a significant improvement in average performance, our results also show strong stability.
Specifically, our MTPL method exhibits lower variance than all four baseline approaches in 24 out of 40 evaluations across the ten tasks, indicating that our method enhances both performance and stability.
However, we observe higher variance in certain tasks, such as $Point\_mass\_easy$ and $Hopper\_hop$, compared to the baselines. 
This higher variance is due to extreme values where our method performs exceptionally well.
This indicates a potential trade-off between high performance and consistency across tasks, rather than a lack of stability in our method.
Furthermore, MTPL enables the baselines to achieve performance levels that approach or even exceed those of SAC, which uses the ground truth reward, in four tasks (Task ID 2, 4, 6, and 10). 
Our MTPL method consistently improves performance across all four baseline methods, demonstrating its effectiveness in enhancing feedback efficiency (Q1) in PBRL. 
We not only outperform the original baselines but also address the challenge posed by the limited number of explicit preferences available for guiding agent behavior. 
\subsection{ABLATION STUDIES}

    \begin{figure*}[htbp] 
        \centering
        \subfigure[PEBBLE]{\includegraphics[width=0.245\textwidth]{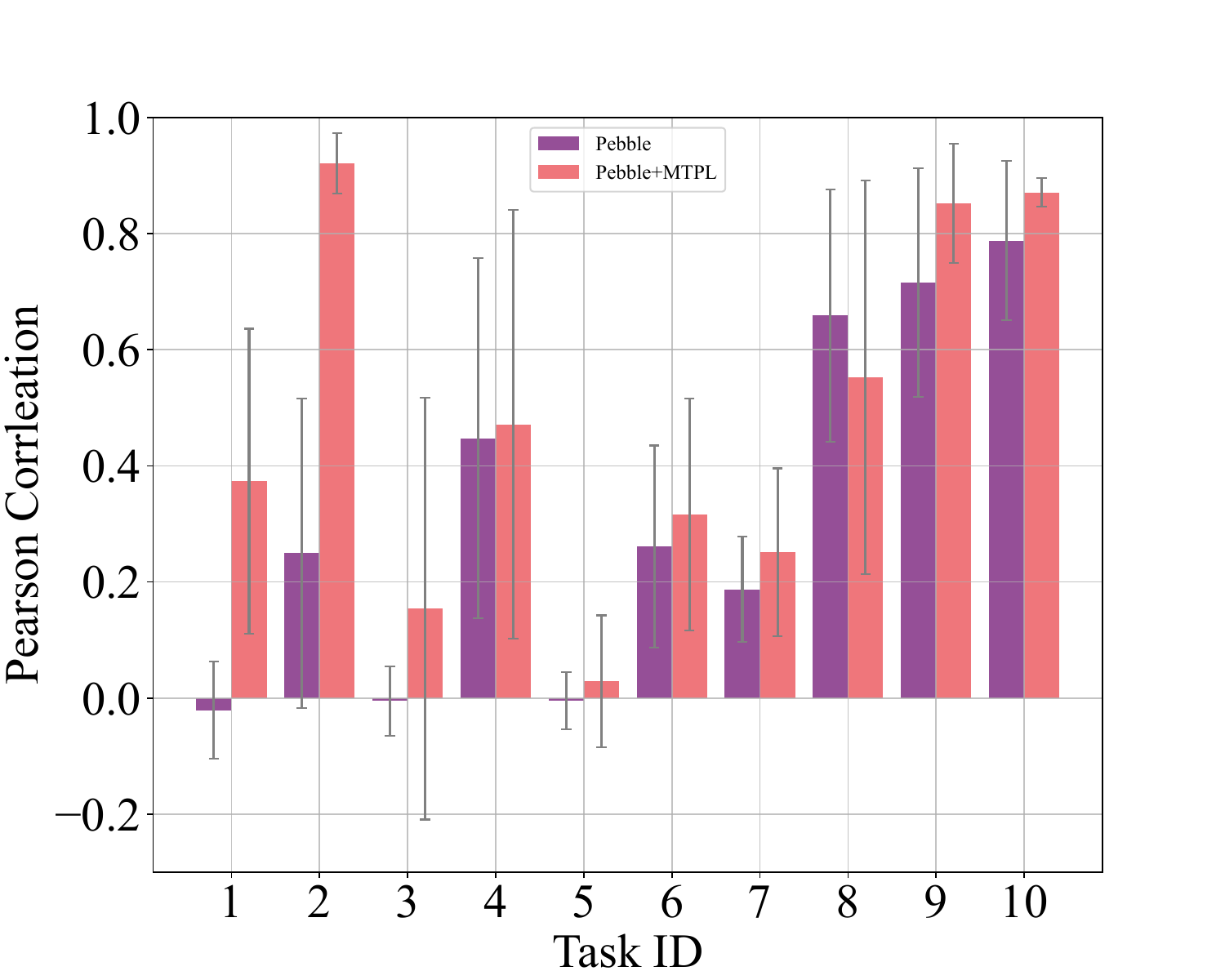}}
        \hfill
        \subfigure[RUNE]{\includegraphics[width=0.245\textwidth]{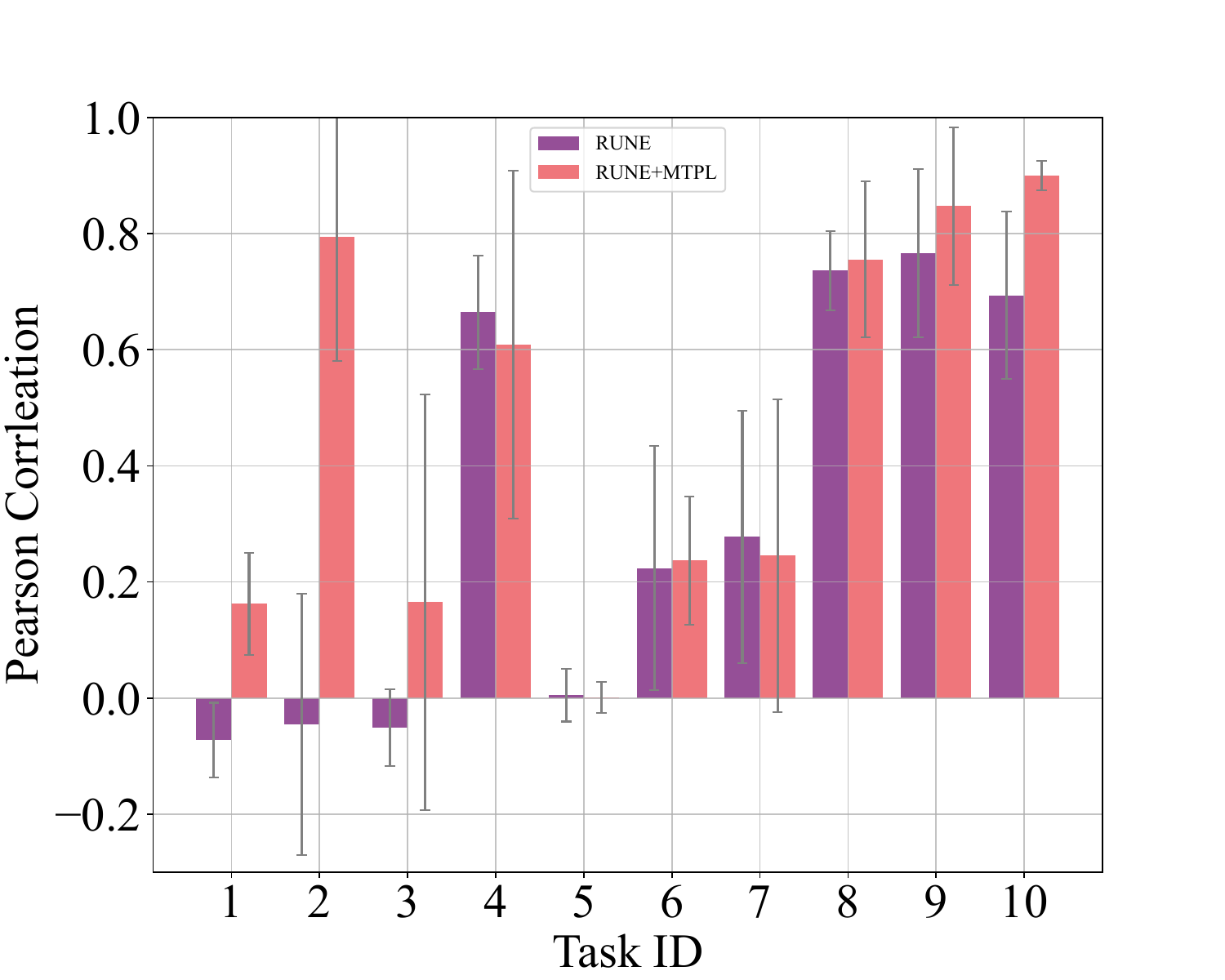}}
        \hfill
        \subfigure[MRN]{\includegraphics[width=0.245\textwidth]{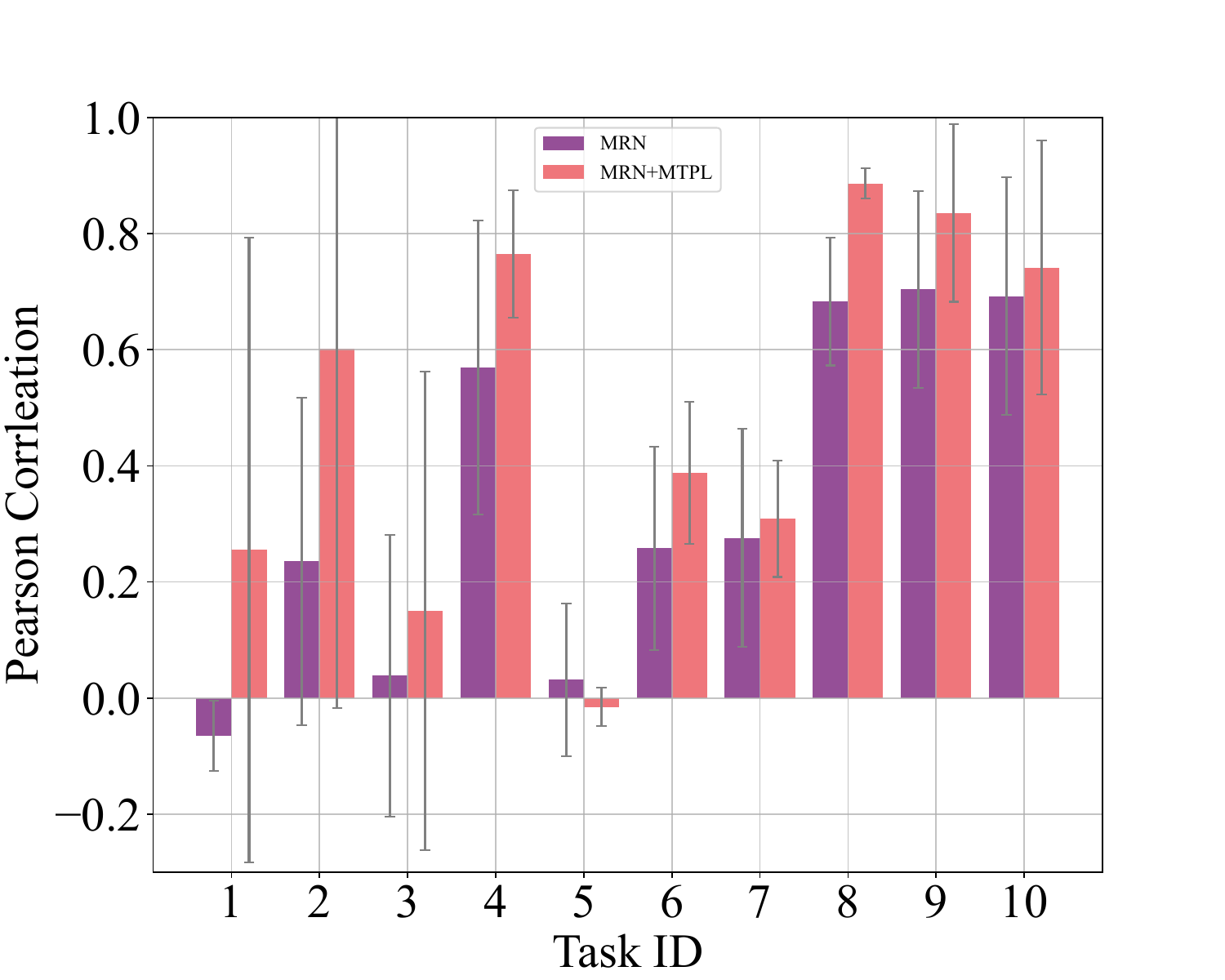}}
        \hfill
        \subfigure[SeqRank]{\includegraphics[width=0.245\textwidth]{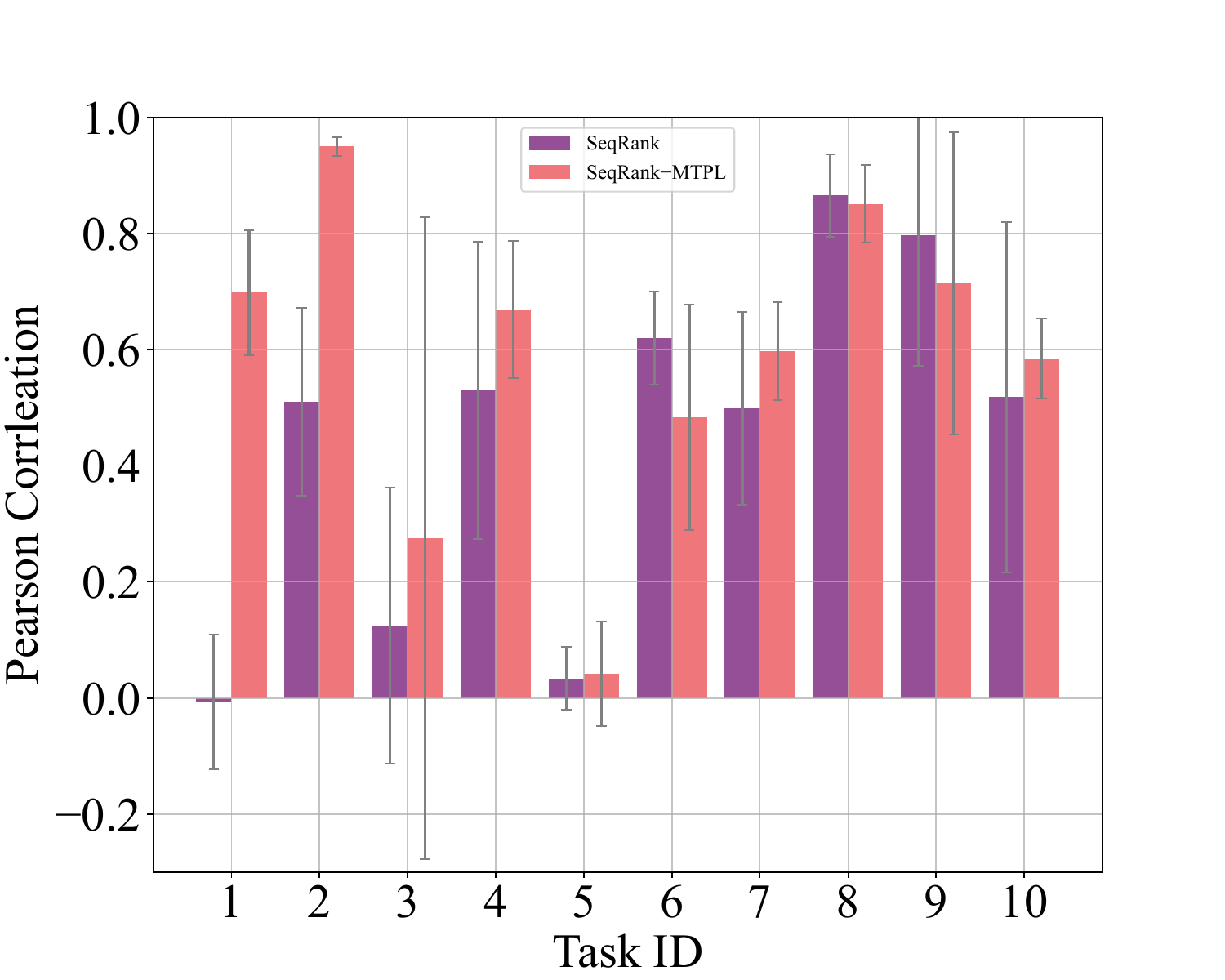}}
        \caption{Comparison of Pearson Correlation Coefficients between Learned and True Reward Functions across 10 Tasks.
                This figure compares the Pearson correlation coefficients between the learned reward function outputs and 
                the true reward function outputs for four baseline methods and the proposed method using MTPL. 
                Each bar represents the average correlation coefficient over five independent runs, and 
                the gray lines indicate the standard deviation.}
        \label{fig:four_subfigures}
    \end{figure*}
\textbf{Effectiveness of Improved Reward Function Learning (Q2).} 
To evaluate how well MTPL captures human feedback, we assess the correlation between the learned reward function and the predefined (ground-truth) reward function.
In our experiments, we use a simulated teacher to replace human input, and the predefined reward function serves as an approximation of human understanding of the task. 
In Fig. \ref{fig:four_subfigures}, we compare the reward learning performance of PEBBLE, RUNE, MRN, and SeqRank, both with and without MTPL. Our analysis focuses on the Pearson correlation between the learned reward function and the predefined reward function. 
This analysis quantifies the linear relationship between the learned and predefined reward functions, providing insights into their alignment.
In our analysis of 40 results, only 7 exhibit lower average correlation coefficients with our method.
Meanwhile, most tasks demonstrate lower standard deviations, highlighting the effectiveness of our approach in enhancing performance across most scenarios.
While our analysis of the Pearson correlation indicates a strong alignment between the learned reward function and the predefined reward function, further studies could enhance our understanding of the underlying mechanisms and validate the robustness of our MTPL method.
This exploration is essential for addressing our second research question (Q2) regarding the understanding of human feedback.
    \begin{figure*}[htbp] 
        \centering
        \subfigure[PEBBLE]{\includegraphics[width=0.24\textwidth]{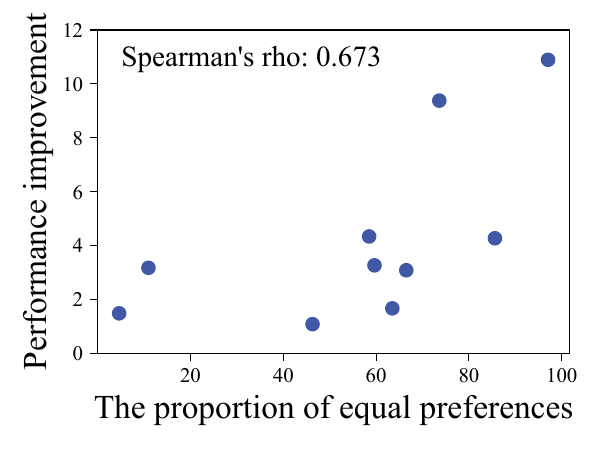}}
        \hfill
        \subfigure[RUNE]{\includegraphics[width=0.24\textwidth]{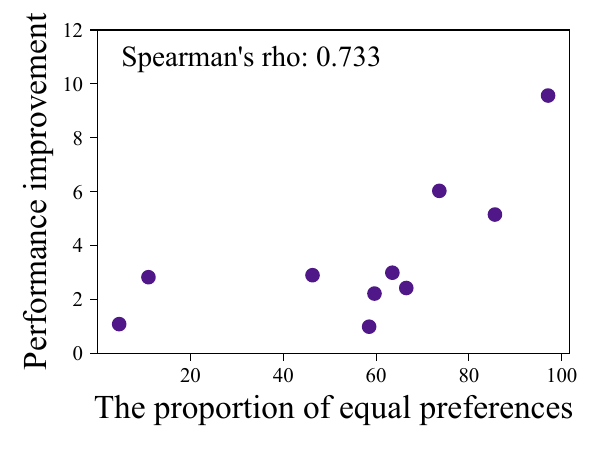}}
        \hfill
        \subfigure[MRN]{\includegraphics[width=0.24\textwidth]{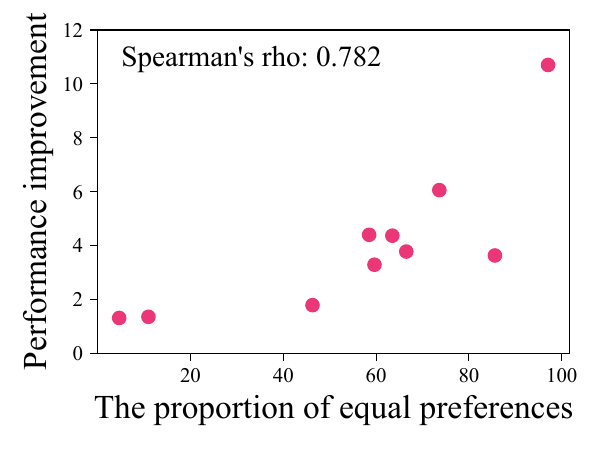}}
        \hfill
        \subfigure[SeqRank]{\includegraphics[width=0.24\textwidth]{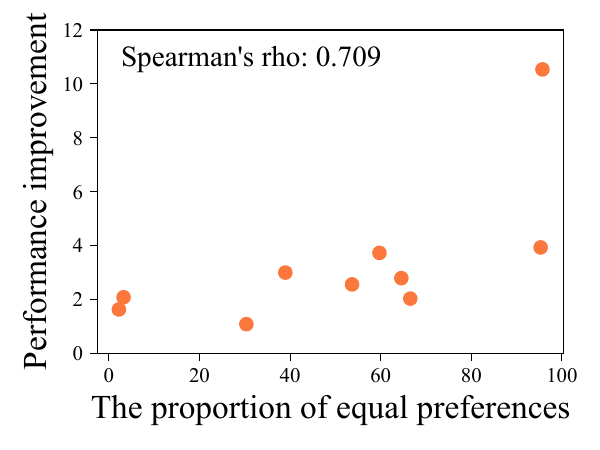}}
        \caption{Spearman Correlation Analysis of Performance Improvement and the Proportion of Equal Preference Feedback.
        This figure presents the Spearman correlation analysis between performance improvement and the proportion of equal preference feedback after applying MTPL to four baseline methods across 10 tasks. To display extreme values, the y-axis has been logarithmically scaled.
        }
                \label{Gin_Corr}
    \end{figure*}
    \begin{figure*}[htbp]
        \centering
        \subfigure[Effects of $\alpha^{Equal}$]{\includegraphics[width=0.3\textwidth]{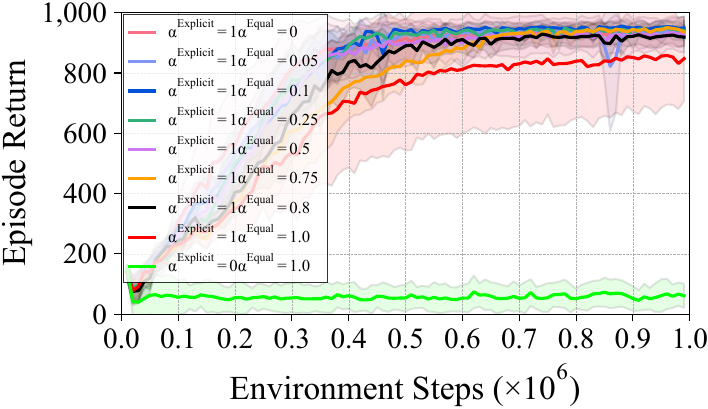}}
        \hfill
        \subfigure[Effects of $\alpha^{Equal}$]{\includegraphics[width=0.3\textwidth]{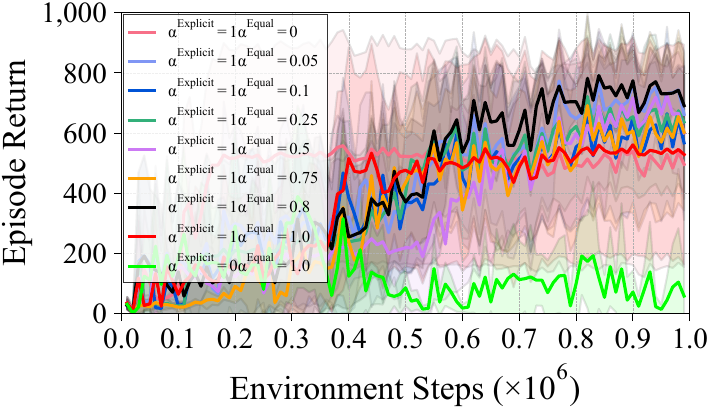}}
        \hfill
        \subfigure[Effects of SimTeacher $\alpha$]{\includegraphics[width=0.3\textwidth]{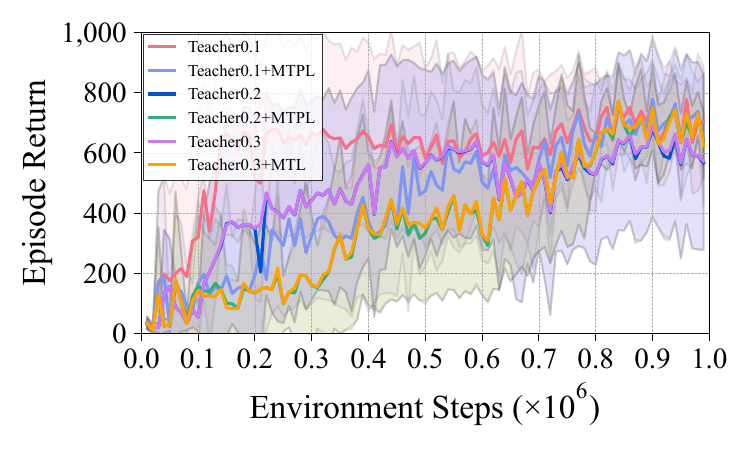}}
        \hfill
        \caption{Hyperparameter Analysis of MTPL. Figure a and Figure b analyze the performance of different parameters $\alpha^{Equal}$ 
        on tasks $Walker\_walk$ and $Pendulum\_swingup$, while Figure c examines the impact of the 
        SimTeacher parameter $\alpha$ on performance.}
        \label{fig_ablation}
    \end{figure*}
\\
\textbf{Correlation Between Performance Improvement and Equal Preferences (Q3).}
Previous methods largely ignored equal preference information, whereas our MTPL method fully utilizes this data. 
As illustrated in Fig. \ref{Gin_Corr}, we analyze the relationship between the proportion of equal preferences across different tasks and the performance improvements achieved with MTPL. 
We employ Spearman correlation analysis, which effectively assesses relationships without assuming normality, is robust to outliers, and captures monotonic relationships.
The results reveal a significant positive correlation, indicating that equal preferences substantially contribute to performance enhancement. 
This suggests that leveraging equal preferences provides valuable insights to improve task performance(Q3), particularly in complex scenarios.
\\
\textbf{Impact of Hyperparameters on MTPL Performance (Q3).}
We further analyze the impact of hyperparameters on MTPL performance within the SOTA baseline MRN \cite{4-MRnet}. 
Specifically, we investigate the effect of $\alpha^{Equal}$ on the tasks $Walker\_walk$ and $Pendulum\_swingup$. These tasks are selected to investigate how variations in hyperparameters influence MTPL performance under both high and low equal preference proportions.
As shown in Fig. \ref{fig_ablation} (a) and (b), our findings indicate that while learning solely from equal preferences is less effective than learning from explicit preferences, it still positively contributes to performance improvements. 
This supports our third research question (Q3), demonstrating that learning from equal preferences provides meaningful insights for task completion.
Therefore, we set $\alpha^{explicit}$ to 1 and primarily adjust $\alpha^{Equal}$. 
We find that learning from both types of preferences simultaneously yields the best results, indicating that MTPL is robust to $\alpha^{Equal}$. 
However, we observe a slight decline in performance when $\alpha^{Equal} = 1$.
While we acknowledge that the convergence speed of MTPL may be slower compared to learning from any single task, we argue that achieving optimal performance with the same number of human preferences is crucial for PBRL methods. 
In Fig. \ref{fig_ablation} (c), we explore the impact of the parameter $\alpha$ of SimTeacher on performance in the $Pendulum\_swingup$ task. 
In our main experiments, we set $\alpha = 0.1$ and observe that as $\alpha$ increases, leading to more ambiguous judgments favoring equal preferences, our method exhibits a more pronounced improvement relative to the baseline. 
This finding underscores that our approach is not limited to a specific $\alpha$ value, highlighting its strong generalization capabilities.

\section{CONCLUSIONS}\label{sec:Con}
Our research introduces Multi-Type Preference Learning (MTPL) as a novel method that enhances feedback efficiency by learning from both explicit and equal preferences. 
The significant performance improvements observed across various tasks indicate that MTPL effectively 
addresses the limitations of prior methods in utilizing equal preferences.
In the future, we anticipate that MTPL will provide an innovative approach for fine-tuning large language models, facilitating better alignment with human values.
This advancement has the potential to drive progress in robotics and automation, particularly in enhancing natural interactions between intelligent agents and humans.

\bibliographystyle{IEEEtran}
\bibliography{IEEEabrv,6ref}

\begin{thebibliography}{10}
\providecommand{\url}[1]{#1}
\csname url@rmstyle\endcsname
\providecommand{\newblock}{\relax}
\providecommand{\bibinfo}[2]{#2}
\providecommand\BIBentrySTDinterwordspacing{\spaceskip=0pt\relax}
\providecommand\BIBentryALTinterwordstretchfactor{4}
\providecommand\BIBentryALTinterwordspacing{\spaceskip=\fontdimen2\font plus
\BIBentryALTinterwordstretchfactor\fontdimen3\font minus
  \fontdimen4\font\relax}
\providecommand\BIBforeignlanguage[2]{{%
\expandafter\ifx\csname l@#1\endcsname\relax
\typeout{** WARNING: IEEEtran.bst: No hyphenation pattern has been}%
\typeout{** loaded for the language `#1'. Using the pattern for}%
\typeout{** the default language instead.}%
\else
\language=\csname l@#1\endcsname
\fi
#2}}

\bibitem{DRL-2022}
X.~Wang, S.~Wang, X.~Liang, D.~Zhao, J.~Huang, X.~Xu, B.~Dai, and Q.~Miao,
  ``Deep reinforcement learning: a survey,'' \emph{IEEE Transactions on Neural
  Networks and Learning Systems}, 2022.

\bibitem{MDP-AI}
R.~S. Sutton, D.~Precup, and S.~Singh, ``Between mdps and semi-mdps: A
  framework for temporal abstraction in reinforcement learning,''
  \emph{Artificial intelligence}, vol. 112, no. 1-2, pp. 181--211, 1999.

\bibitem{RL-1961}
M.~Minsky, ``Steps toward artificial intelligence,'' \emph{Proceedings of the
  IRE}, vol.~49, no.~1, pp. 8--30, 1961.

\bibitem{EPFL-Magnetic-control}
J.~Degrave, F.~Felici, J.~Buchli, M.~Neunert, B.~Tracey, F.~Carpanese,
  T.~Ewalds, R.~Hafner, A.~Abdolmaleki, D.~de~Las~Casas, \emph{et~al.},
  ``Magnetic control of tokamak plasmas through deep reinforcement learning,''
  \emph{Nature}, vol. 602, no. 7897, pp. 414--419, 2022.

\bibitem{ETH-swift}
E.~Kaufmann, L.~Bauersfeld, A.~Loquercio, M.~M{\"u}ller, V.~Koltun, and
  D.~Scaramuzza, ``Champion-level drone racing using deep reinforcement
  learning,'' \emph{Nature}, vol. 620, no. 7976, pp. 982--987, 2023.

\bibitem{DPRL-IJRR}
E.~B{\i}y{\i}k, D.~P. Losey, M.~Palan, N.~C. Landolfi, G.~Shevchuk, and
  D.~Sadigh, ``Learning reward functions from diverse sources of human
  feedback: Optimally integrating demonstrations and preferences,'' \emph{The
  International Journal of Robotics Research}, vol.~41, no.~1, pp. 45--67,
  2022.

\bibitem{26-icra}
D.~Quillen, E.~Jang, O.~Nachum, C.~Finn, J.~Ibarz, and S.~Levine, ``Deep
  reinforcement learning for vision-based robotic grasping: A simulated
  comparative evaluation of off-policy methods,'' in \emph{2018 IEEE
  international conference on robotics and automation (ICRA)}.\hskip 1em plus
  0.5em minus 0.4em\relax IEEE, 2018, pp. 6284--6291.

\bibitem{27-icra}
G.~Kahn, A.~Villaflor, B.~Ding, P.~Abbeel, and S.~Levine, ``Self-supervised
  deep reinforcement learning with generalized computation graphs for robot
  navigation,'' in \emph{2018 IEEE international conference on robotics and
  automation (ICRA)}.\hskip 1em plus 0.5em minus 0.4em\relax IEEE, 2018, pp.
  5129--5136.

\bibitem{0-2017survey}
C.~Wirth, R.~Akrour, G.~Neumann, and J.~F{\"u}rnkranz, ``A survey of
  preference-based reinforcement learning methods,'' \emph{Journal of Machine
  Learning Research}, vol.~18, no. 136, pp. 1--46, 2017.

\bibitem{23-ICML}
W.~Xiong, H.~Dong, C.~Ye, Z.~Wang, H.~Zhong, H.~Ji, N.~Jiang, and T.~Zhang,
  ``Iterative preference learning from human feedback: Bridging theory and
  practice for rlhf under kl-constraint,'' in \emph{Forty-first International
  Conference on Machine Learning}, 2024.

\bibitem{mis-specified-1}
D.~Hadfield-Menell, S.~Milli, P.~Abbeel, S.~J. Russell, and A.~Dragan,
  ``Inverse reward design,'' \emph{NeurIPS}, vol.~30, 2017.

\bibitem{mis-specified-2}
A.~Turner, N.~Ratzlaff, and P.~Tadepalli, ``Avoiding side effects in complex
  environments,'' \emph{NeurIPS}, vol.~33, pp. 21\,406--21\,415, 2020.

\bibitem{0-DRLPB}
P.~F. Christiano, J.~Leike, T.~Brown, M.~Martic, S.~Legg, and D.~Amodei, ``Deep
  reinforcement learning from human preferences,'' \emph{NeurIPS}, vol.~30,
  2017.

\bibitem{19-rlhf}
S.~Casper, X.~Davies, C.~Shi, T.~K. Gilbert, J.~Scheurer, J.~Rando,
  R.~Freedman, T.~Korbak, D.~Lindner, P.~Freire, \emph{et~al.}, ``Open problems
  and fundamental limitations of reinforcement learning from human feedback,''
  \emph{arXiv preprint arXiv:2307.15217}, 2023.

\bibitem{25-rlhfsurvey}
Z.~Wang, B.~Bi, S.~K. Pentyala, K.~Ramnath, S.~Chaudhuri, S.~Mehrotra, X.-B.
  Mao, S.~Asur, \emph{et~al.}, ``A comprehensive survey of llm alignment
  techniques: Rlhf, rlaif, ppo, dpo and more,'' \emph{arXiv preprint
  arXiv:2407.16216}, 2024.

\bibitem{28-2024human}
C.~O. Retzlaff, S.~Das, C.~Wayllace, P.~Mousavi, M.~Afshari, T.~Yang,
  A.~Saranti, A.~Angerschmid, M.~E. Taylor, and A.~Holzinger,
  ``Human-in-the-loop reinforcement learning: A survey and position on
  requirements, challenges, and opportunities,'' \emph{Journal of Artificial
  Intelligence Research}, vol.~79, pp. 359--415, 2024.

\bibitem{1-PEBBLE}
K.~Lee, L.~Smith, and P.~Abbeel, ``Pebble: Feedback-efficient interactive
  reinforcement learning via relabeling experience and unsupervised
  pre-training,'' in \emph{ICML}, 2021.

\bibitem{2-RUNE}
X.~Liang, K.~Shu, K.~Lee, and P.~Abbeel, ``Reward uncertainty for exploration
  in preference-based reinforcement learning,'' in \emph{ICLR}, 2021.

\bibitem{14-loss}
M.~Verma, S.~Bhambri, and S.~Kambhampati, ``Exploiting unlabeled data for
  feedback efficient human preference based reinforcement learning,''
  \emph{arXiv preprint arXiv:2302.08738}, 2023.

\bibitem{3-SURF}
J.~Park, Y.~Seo, J.~Shin, H.~Lee, P.~Abbeel, and K.~Lee, ``{SURF}:
  Semi-supervised reward learning with data augmentation for feedback-efficient
  preference-based reinforcement learning,'' in \emph{ICLR}, 2022.

\bibitem{6-FSHITLRL}
D.~J. Hejna~III and D.~Sadigh, ``Few-shot preference learning for
  human-in-the-loop rl,'' in \emph{CoRL}.\hskip 1em plus 0.5em minus
  0.4em\relax PMLR, 2023, pp. 2014--2025.

\bibitem{4-MRnet}
R.~Liu, F.~Bai, Y.~Du, and Y.~Yang, ``Meta-reward-net: Implicitly
  differentiable reward learning for preference-based reinforcement learning,''
  in \emph{NeurIPS}, vol.~35, 2022, pp. 22\,270--22\,284.

\bibitem{11-Seqrank}
M.~Hwang, G.~Lee, H.~Kee, C.~W. Kim, K.~Lee, and S.~Oh, ``Sequential preference
  ranking for efficient reinforcement learning from human feedback,''
  \emph{Advances in Neural Information Processing Systems}, vol.~36, 2024.

\bibitem{18-B-pre}
K.~Lee, L.~Smith, A.~Dragan, and P.~Abbeel, ``B-pref: Benchmarking
  preference-based reinforcement learning,'' in \emph{Thirty-fifth Conference
  on Neural Information Processing Systems Datasets and Benchmarks Track},
  2021.

\bibitem{17-DMC}
Y.~Tassa, Y.~Doron, A.~Muldal, T.~Erez, Y.~Li, D.~d.~L. Casas, D.~Budden,
  A.~Abdolmaleki, J.~Merel, A.~Lefrancq, \emph{et~al.}, ``Deepmind control
  suite,'' \emph{arXiv preprint arXiv:1801.00690}, 2018.

\bibitem{12-MTLsurvey}
J.~Yu, Y.~Dai, X.~Liu, J.~Huang, Y.~Shen, K.~Zhang, R.~Zhou, E.~Adhikarla,
  W.~Ye, Y.~Liu, \emph{et~al.}, ``Unleashing the power of multi-task learning:
  A comprehensive survey spanning traditional, deep, and pretrained foundation
  model eras,'' \emph{arXiv preprint arXiv:2404.18961}, 2024.

\bibitem{28-IROS}
D.~Marta, S.~Holk, C.~Pek, J.~Tumova, and I.~Leite, ``Variquery: Vae
  segment-based active learning for query selection in preference-based
  reinforcement learning,'' in \emph{2023 IEEE/RSJ International Conference on
  Intelligent Robots and Systems (IROS)}.\hskip 1em plus 0.5em minus
  0.4em\relax IEEE, 2023, pp. 7878--7885.

\bibitem{34-FTB}
Z.~Zhang, Y.~Sun, J.~Ye, T.-S. Liu, J.~Zhang, and Y.~Yu, ``Flow to better:
  Offline preference-based reinforcement learning via preferred trajectory
  generation,'' in \emph{The Twelfth International Conference on Learning
  Representations}, 2023.

\bibitem{13-SAC}
T.~Haarnoja, A.~Zhou, P.~Abbeel, and S.~Levine, ``Soft actor-critic: Off-policy
  maximum entropy deep reinforcement learning with a stochastic actor,'' in
  \emph{ICML}, 2018, pp. 1861--1870.

\bibitem{5-LSI}
G.~Zhang and H.~Kashima, ``Learning state importance for preference-based
  reinforcement learning,'' \emph{Machine Learning}, pp. 1--17, 2023.

\bibitem{10-iclr}
M.~Verma and K.~Metcalf, ``Hindsight priors for reward learning from human
  preferences,'' in \emph{The Twelfth International Conference on Learning
  Representations, {ICLR} 2024, Vienna, Austria, May 7-11, 2024}.\hskip 1em
  plus 0.5em minus 0.4em\relax OpenReview.net, 2024.

\bibitem{20-AAAI24}
D.~White, M.~Wu, E.~Novoseller, V.~J. Lawhern, N.~Waytowich, and Y.~Cao,
  ``Rating-based reinforcement learning,'' in \emph{Proceedings of the AAAI
  Conference on Artificial Intelligence}, vol.~38, no.~9, 2024, pp.
  10\,207--10\,215.

\bibitem{33-PT}
C.~Kim, J.~Park, J.~Shin, H.~Lee, P.~Abbeel, and K.~Lee, ``Preference
  transformer: Modeling human preferences using transformers for rl,'' in
  \emph{Eleventh International Conference on Learning Representations, ICLR
  2023}.\hskip 1em plus 0.5em minus 0.4em\relax International Conference on
  Learning Representations, 2023.

\bibitem{22-AAAI2024}
T.~Zhu, Y.~Qiu, H.~Zhou, and J.~Li, ``Decoding global preferences: Temporal and
  cooperative dependency modeling in multi-agent preference-based reinforcement
  learning,'' in \emph{Proceedings of the AAAI Conference on Artificial
  Intelligence}, vol.~38, no.~15, 2024, pp. 17\,202--17\,210.

\bibitem{29-MTLsurvey}
Y.~Zhang and Q.~Yang, ``A survey on multi-task learning,'' \emph{IEEE
  transactions on knowledge and data engineering}, vol.~34, no.~12, pp.
  5586--5609, 2021.

\bibitem{14-aaai}
F.~Bai, H.~Zhang, T.~Tao, Z.~Wu, Y.~Wang, and B.~Xu, ``Picor: Multi-task deep
  reinforcement learning with policy correction,'' in \emph{Proceedings of the
  AAAI Conference on Artificial Intelligence}, vol.~37, no.~6, 2023, pp.
  6728--6736.

\bibitem{32-MTLrl}
S.~Sodhani, A.~Zhang, and J.~Pineau, ``Multi-task reinforcement learning with
  context-based representations,'' in \emph{International Conference on Machine
  Learning}.\hskip 1em plus 0.5em minus 0.4em\relax PMLR, 2021, pp. 9767--9779.

\bibitem{15-nips}
H.~Li, K.~Wu, C.~Zheng, Y.~Xiao, H.~Wang, Z.~Geng, F.~Feng, X.~He, and P.~Wu,
  ``Removing hidden confounding in recommendation: a unified multi-task
  learning approach,'' \emph{Advances in Neural Information Processing
  Systems}, vol.~36, 2024.

\bibitem{30-MTLre}
M.~Gao, J.-Y. Li, C.-H. Chen, Y.~Li, J.~Zhang, and Z.-H. Zhan, ``Enhanced
  multi-task learning and knowledge graph-based recommender system,''
  \emph{IEEE Transactions on Knowledge and Data Engineering}, vol.~35, no.~10,
  pp. 10\,281--10\,294, 2023.

\bibitem{16-nmi}
S.~Allenspach, J.~A. Hiss, and G.~Schneider, ``Neural multi-task learning in
  drug design,'' \emph{Nature Machine Intelligence}, vol.~6, no.~2, pp.
  124--137, 2024.

\bibitem{31-MTLdd}
S.~Lin, C.~Shi, and J.~Chen, ``Generalizeddta: combining pre-training and
  multi-task learning to predict drug-target binding affinity for unknown drug
  discovery,'' \emph{BMC bioinformatics}, vol.~23, no.~1, p. 367, 2022.

\end{thebibliography}


\addtolength{\textheight}{-12cm}   



\section*{APPENDIX}
\begin{figure*}[htbp]
    \centering
    \subfigure[Point mass easy]{\includegraphics[width=0.48\textwidth]{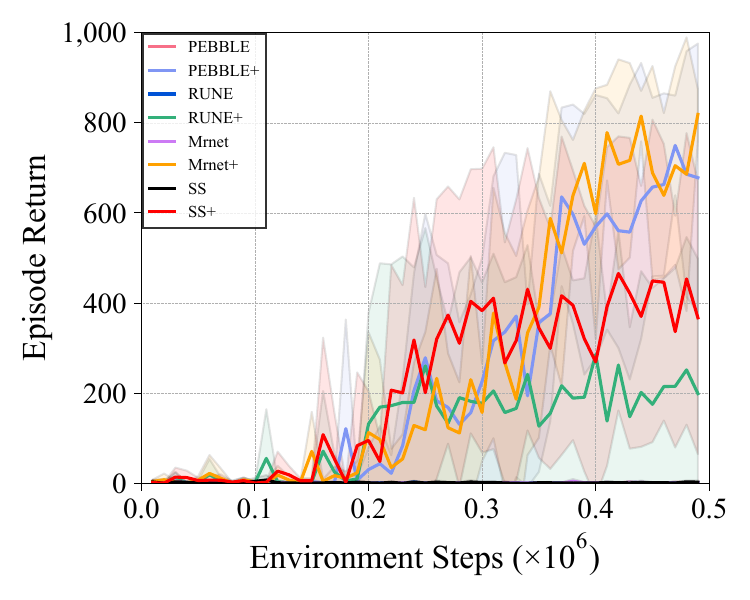}} \hfill
    \subfigure[Hopper Hop]{\includegraphics[width=0.48\textwidth]{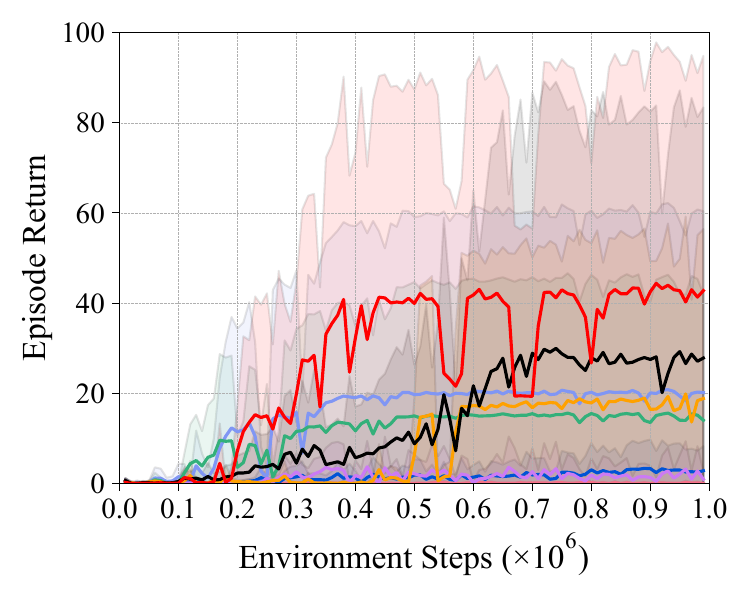}} \hfill
        \vspace{0.5cm} 
    \subfigure[Cartpole balance sparse]{\includegraphics[width=0.23\textwidth]{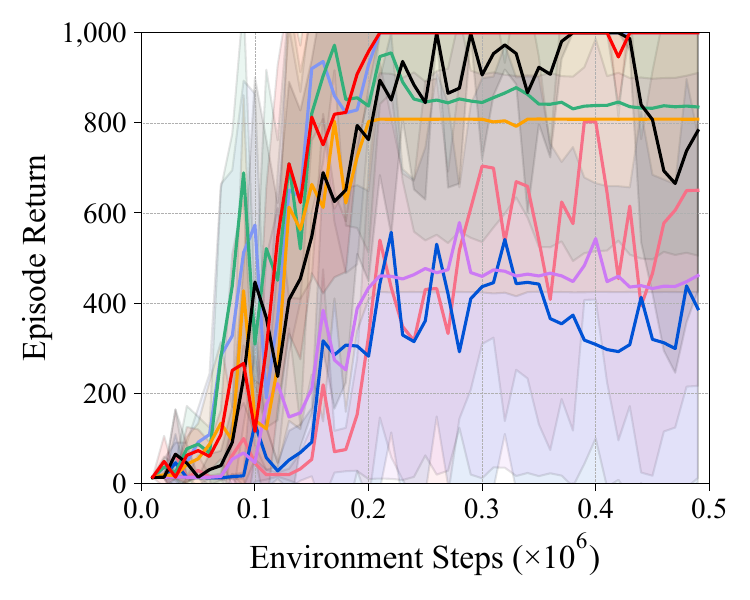}} \hfill
    \subfigure[Ball in cup catch]{\includegraphics[width=0.23\textwidth]{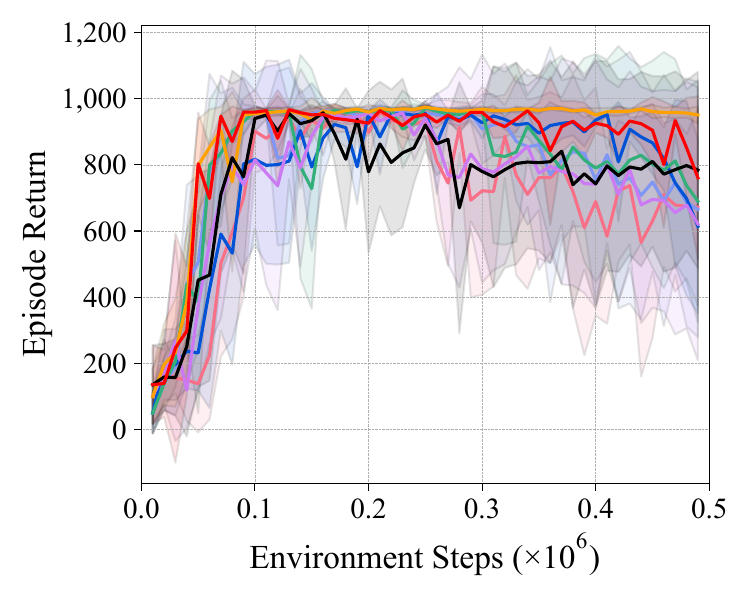}} 
    \subfigure[Reacher easy]{\includegraphics[width=0.23\textwidth]{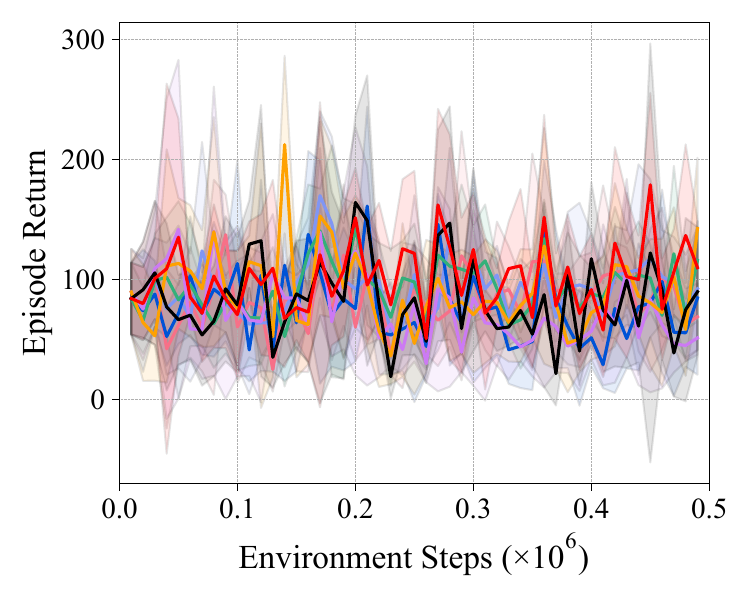}} \hfill
    \subfigure[Finger turn Hard]{\includegraphics[width=0.23\textwidth]{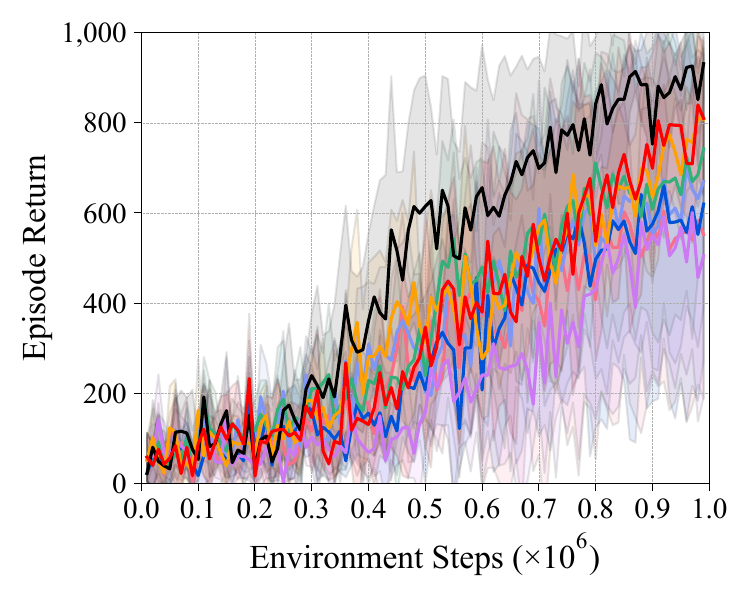}} \hfill
    \vspace{0.5cm} 


    \subfigure[Finger turn Easy]{\includegraphics[width=0.23\textwidth]{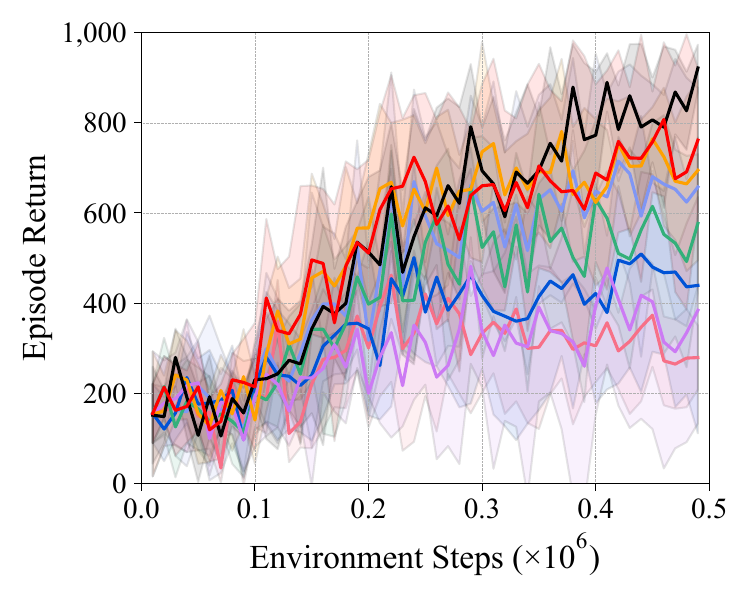}} \hfill
    \subfigure[Pendulum Swingup]{\includegraphics[width=0.23\textwidth]{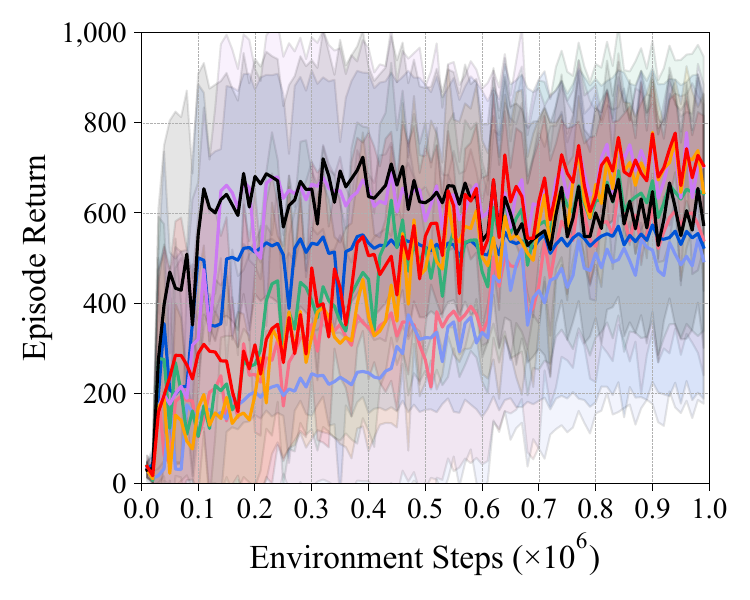}}
    \subfigure[Quadruped Walk]{\includegraphics[width=0.23\textwidth]{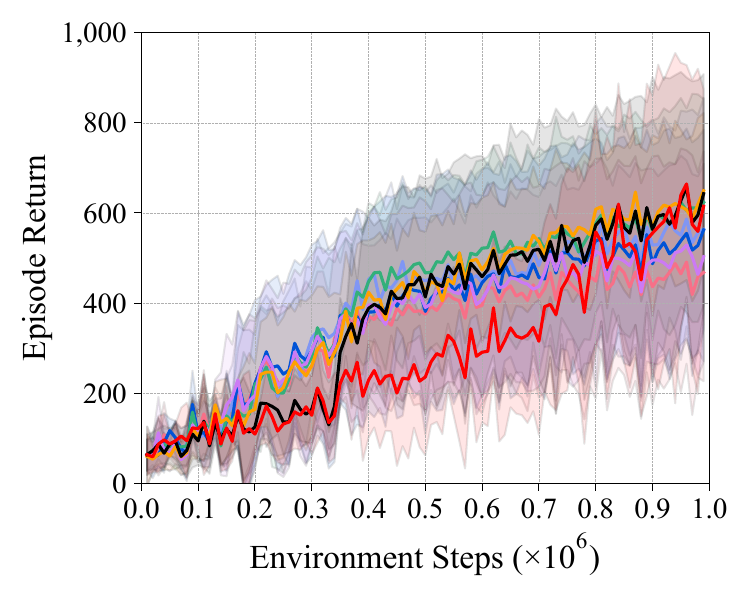}} \hfill
    \subfigure[Walker Walk]{\includegraphics[width=0.23\textwidth]{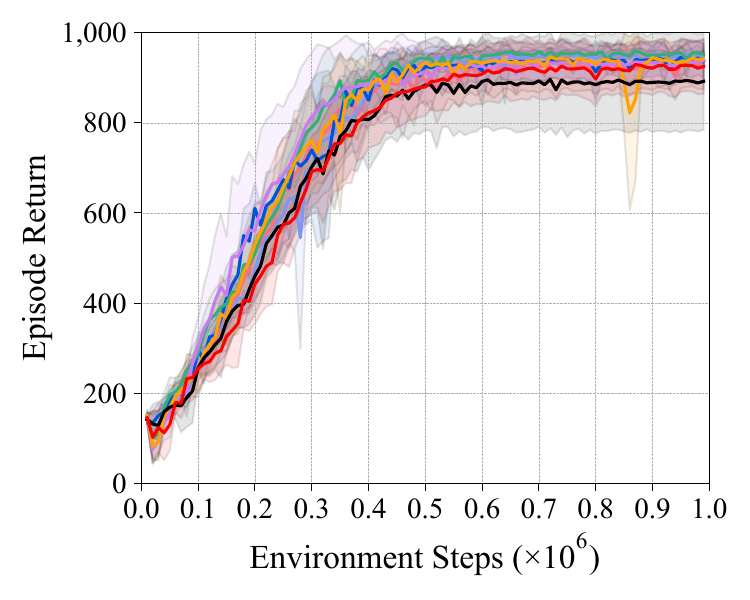}}

    \caption{The learning curves of four baselines compared to MTPL across ten tasks.}
    \label{fig:all_images}
\end{figure*}


\end{document}